\documentclass[conference,a4paper]{IEEEtran}
\IEEEoverridecommandlockouts

\usepackage[dvips]{graphicx}
\usepackage{subfigure}
\usepackage{eclbkbox}
\usepackage{indent}

\makeatletter
\def\tbcaption{\def\@captype{table}\caption}
\def\figcaption{\def\@captype{figure}\caption}

\ifCLASSINFOpdf
\else
\fi

\begin{document}

\title{Clustering and Retrieval Method of Immunological Memory Cell in Clonal Selection Algorithm
\thanks{\copyright 2012 IEEE. Personal use of this material is permitted. Permission from IEEE must be obtained for all other uses, in any current or future media, including reprinting/republishing this material for advertising or promotional purposes, creating new collective works, for resale or redistribution to servers or lists, or reuse of any copyrighted component of this work in other works.}
}

\author{\IEEEauthorblockN{Takumi Ichimura}
\IEEEauthorblockA{Faculty of Management and Information Systems,\\
Prefectural University of Hiroshima\\
1-1-71, Ujina-Higashi, Minami-ku,\\
Hiroshima, 734-8558, Japan\\
Email: ichimura@pu-hiroshima.ac.jp}

\and
\IEEEauthorblockN{Shin Kamada}
\IEEEauthorblockA{Graduate School of Comprehensive Scientific Research\\
Prefectural University of Hiroshima,\\
1-1-71, Ujina-Higashi, Minami-ku,\\
Hiroshima, 734-8558, Japan\\
Email: shinkamada46@gmail.com}
}

\maketitle

\begin{abstract}
The clonal selection principle explains the basic features of an adaptive immune response to a antigenic stimulus. It established the idea that only those cells that recognize the antigens are selected to proliferate and differentiate. This paper explains a computational implementation of the clonal selection principle that explicitly takes into account the affinity maturation of the immune response. Antibodies generated by the clonal selection algorithm are clustered in some categories according to the affinity maturation, so that immunological memory cells which respond to the specified pathogen are created. Experimental results to classify the medical database of Coronary Heart Disease databases are reported. For the dataset, our proposed method shows the 99.6\% classification capability of training data.
\end{abstract}

\IEEEpeerreviewmaketitle

\section{Introduction}
\label{sec:Introduction}
In a few decades, the area of artificial immune system (AIS) has been an ever-increasing interested in not only theoretical works but applications in pattern recognition, network security, and optimization \cite{Castro1},\cite{Castro2},\cite{Castro3},\cite{Hunt},\cite{Dasgupta},\cite{Hofmeyr},\cite{Ishida}. AIS uses ideas gleaned from immunology in order to develop adaptive systems capable of performing a wide range of tasks in various research areas.

The biological basis of the Clonal Selection Theory was proposed by Burent \cite{Burnet1},\cite{Burnet2} in 1959. The theory interprets the response of lymphocytes in the face of antigenic stimulus. Only the immune cells with high affinity are selected to proliferate, while those low affinity cells must be efficiently deleted or become anergic. The hypermutation is allowed to improve the affinity of the selected cells to the selective antigens. Receptor Editing as a mechanism of immune cell tolerance is reported \cite{Verkoczy},\cite{Pelanda},\cite{Longo}.

Gao indicated the complementary roles of somatic hypermutation (HM) and receptor editing (RE) and presented a novel clonal selection algorithm called RECSA model by incorporating the Receptor Editing method \cite{Gao}. In \cite{Gao}, they discussed the relationship between HM and RE through utilizing them to solve the TSPs. Because a valid tour in TSP is represented by a permutation of $N$ cities, the number of states to feasible tours is $(N-1)!$ [1] solves the problems for finding an optimal set in the search space consisted of the set of $(N-1)!$ valid tours in TSPs.

A gene structure to classify the Coronary Heart Disease database (CHD\_DB)\cite{Suka-Data} by using RECSA model was developed in \cite{Ichimura2010}. The classification of medical database differs from the TSP, because medical information such as results of biochemical tests and chief complaint is often ambiguous. Therefore, we cannot clearly distinguish the difference between normal and pathological values. Biochemical test values cannot be precisely evaluated by using crisp sets. An output in the database is summed up all the inputs modified by their respective weights and is compared with the corresponding teach signal. If the summation of weighted input is larger than a certain threshold, the antibody respond to the training sample. We consider that the threshold value is important to classify the samples. Therefore, the memory cells are trained to respond to the specified samples according to the threshold values. In order to the effectiveness of the proposed method, we challenge the computational classification results of the databases. The database consists of Train\_A, X, Y, and Z and one testing dataset. Our proposed method has the extraordinary power, since the simulation results shows 99.6\% classification capability of the training dataset, Train\_A.

The remainder of this paper is organized as follows. In Section \ref{sec:CS}, the clonal selection theory will be explained briefly. Section \ref{sec:CSAIM} will explain RECSA model proposed in \cite{Gao}. The response by memory cells will be discussed in Section \ref{sec:IMC}. Experimental results for classification of the medical database will be reported in Section \ref{sec:ExperimentalResults}. In Section \ref{sec:ConclusiveDiscussion}, we give some discussions to conclude this paper.

\section{The Clonal Selection Theory}
\label{sec:CS}
Burnet proposed the clonal selection theory in order to explain the essential features of adaptive immune response \cite{Burnet1}\cite{Burnet2}. The basic idea of this theory interprets the response of lymphocytes in the face of an antigenic stimulus. 

Any molecule that can be recognized by the adaptive immune system is known as antigens ({\it Ag}s).  Some subpopulations of its bone-marrow-derived cells responds by producing antibody ({\it Ab}). {\it Ab}s are molecules attached primarily to the surface of B cells. The aim of B cell is to recognize and bind to {\it Ag}s. Each B cell (B lymphocytes) secretes a single type of {\it Ab}. By binding to these {\it Ab}s and with a second signal from T-helper cell, the {\it Ag} stimulates the B cell to proliferate and mature into terminal {\it Ab} secreting cells called plasma cells. Proliferation of the B cell is a mitotic process whereby the cells divide themselves, creating a set of clones identical to the parent cell. The proliferation rate is directly proportional to the affinity level. That is, the higher affinity level of B cells, the more of them will be readily selected for cloning and cloned in larger numbers.

In addition to proliferating and differentiating into plasma cells, B cells can differentiate into long lived B memory cells. Memory cells circulate through the blood, lymph, and tissues. By exposing to a second antigenic stimulus, they commence to differentiate into plasma cells capable of producing high affinity {\it Ab}s, which are preselected for the specific {\it Ag} that stimulated the primary response.

Gao has been proposed the clonal selection method by incorporating receptor editing which can diversify the repertoire of antigen activated B cells during asexual reproduction \cite{Gao}. The method has two mechanisms of somatic hypermutation and receptor editing \cite{Nussenzweig}\cite{Tonegawa}\cite{Kouskoff}.

\section{Clonal Selection Algorithm with Immunological Memory}
\label{sec:CSAIM}
Clonal Selection Algorithm with Immunological Memory(CSAIM) model has been proposed to introduce an idea of immunological memory into the RECSA model. This section describes the structure of antibody in RECSA model to the medical diagnosis briefly.

\subsection{RECSA Model\cite{Gao}}
The shape-space model aims at quantitatively describing the interactions among {\it Ag}s and {\it Ab}s ({\it Ag}-{\it Ab}) \cite{Perelson1},\cite{Perelson2}. The set of features that characterize a molecule is called its generalized shape. The {\it Ag}-{\it Ab} codification determines their spatial representation and a distance measure is used to calculate the degree of interaction between these molecules. 

The Gao's model \cite{Gao} can be described as follows.
\begin{description}
%\begin{indentation}{0.1zw}{0.1zw}
\begin{indentation}{0.1cm}{0.1cm}

\item[1)] Create an initial pool of $m$ antibodies as candidate solutions$({\it Ab_{1}}, {\it Ab_{2}}, \cdots, {\it Ab_{m}})$.

\item[2)] Compute the affinity of all antibodies: $(D({\it Ab_{1}}), D({\it Ab_{2}}), \cdots, D({\it Ab_{m}}))$. $D()$ means the function to compute the affinity.

\item[3)] Select $n$ best individuals based on their affinities from the $m$ original antibodies. These antibodies will be referred to as the elites.

\item[4)] Sort the $n$ selected elites in $n$ separate and distinct pools in ascending order. They will be referred to as the elite pools.

\item[5)] Clone the elites in the pool with a rate proportional to its fitness. The amount of clone generated for these antibodies is given by Eq.(\ref{eq2-1}).

\begin{equation}
P_{i} = round(\frac{n-i}{n}\times Q )
\label{eq2-1}
\end{equation}
$i$ is the ordinal number of the elite pools, $Q$ is a multiplying factor for determining the scope of the clone and $round()$ is the operator that rounds towards the closest integer. Then, we can obtain $\sum P_{i}$ antibodies as $(({\it Ab_{1,1}}, {\it Ab_{1,2}}, \cdots,$ ${\it Ab_{1,p_{1}}}), \cdots, ({\it Ab_{n,1}}, {\it Ab_{n,2}}, \cdots, {\it Ab_{n,p_{n}}}))$.

\item[6)] Subject the clones in each pool through either hypermutation or receptor editing process. The mutation rates, $P_{hm}$ for hypermutation and $P_{re}$ for receptor editing given by Eq.(\ref{eq2-2}) and Eq.(\ref{eq2-3}), are inversely proportional to the fitness of the parent antibody,
\begin{equation}
P_{hm} = a/D()
\label{eq2-2}
\end{equation}
\begin{equation}
P_{re} = (D()-a)/D(),
\label{eq2-3}
\end{equation}
where $D()$ is the affinity of the current parent antibody and $a$ is an appropriate numerous value.

\item[7)] Determine the fittest individual $B_{i}$ in each elite pool from amongst its mutated clones. The $B_{i}$ is satisfied with the following equation.
\begin{eqnarray}
\nonumber D(B_{i}) &=& max(D({\it Ab_{i,1}}), \cdots, D({\it Ab_{i,p_{i}}})),\\
i&=&1,2,\cdots,n
\label{eq3-4}
\end{eqnarray}

\item[8)] Update the parent antibodies in each elite pool with the fittest individual of the clones and the probability $P({\it Ab_{i}}, \rightarrow B_{i})$ is according to the roles: if $D({\it Ab_{i}}) < D(B_{i})$ then $P=1$, if $D({\it Ab_{1}}) \ge D(B_{1})$ then $P=0$, if $D({\it Ab_{i}}) \ge D(B_{i}), i \ne 1$ then $P=exp(\frac{D(B_{i})-D({\it Ab_{i}})}{\alpha})$.

\item[9)] Replace the worst $c(=\beta n, \beta is arbitrary value.)$ elite pools with new random antibodies once every $t$ generations to introduce diversity and prevent the search from being trapped in local optima. 

\item[10)] Determine if the maximum number of generation $G_{max}$ to evolve is reached. If it is satisfied with this condition, it terminates and returns the best antibody. Otherwise, go to Step 4).
\end{indentation}
\end{description}

\subsubsection{Structure of Antibody to Medical Diagnosis}
\begin{figure}[htbp]
\begin{center}
\includegraphics[scale=0.5]{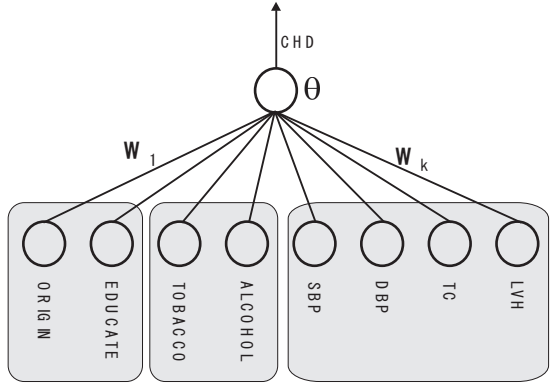}
\caption{The antibody structure}
\label{fig:immune_CHD}
\end{center}
\end{figure}

Fig.\ref{fig:immune_CHD} shows the structure of antibody in the classification problem of Heart Disease Database\cite{Suka-Data}. Each of the data sets consists of ten data items as shown in Table \ref{table:CHDdata}.

\begin{figure}[htbp]
\begin{center}
\includegraphics[scale=0.5]{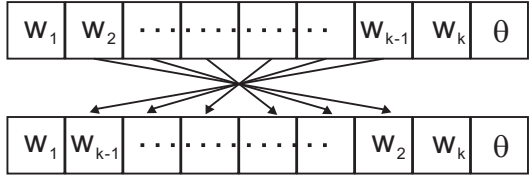}
\caption{RE for $w_2,w_{k-1}$}
\label{fig:network_gene}
\end{center}
\end{figure}

\begin{figure*}[htbp]
\begin{center}
\includegraphics[scale=1.0]{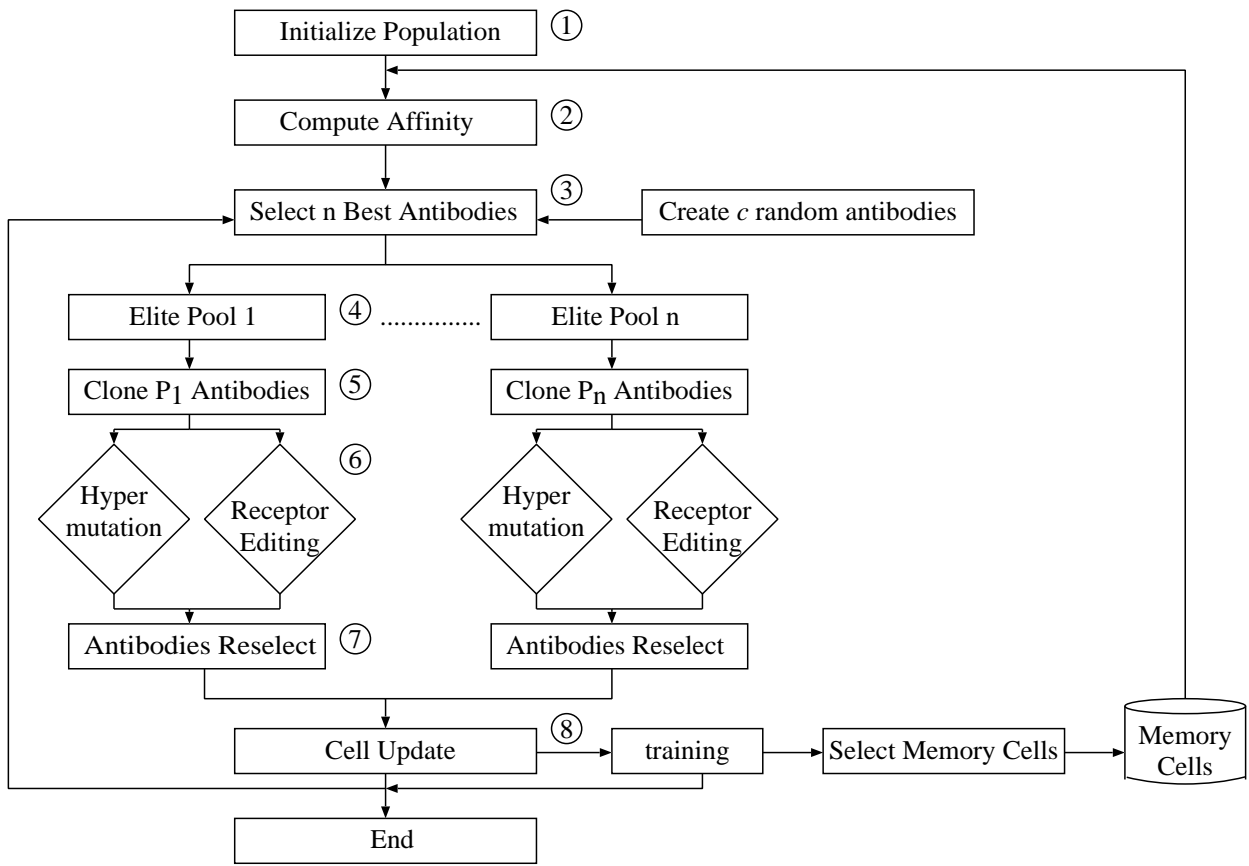}
\caption{A flow of CSAIM model}
\label{fig:CSAIMmodel}
\end{center}
\end{figure*}

\subsubsection{Somatic Hypermutation and Receptor Editing}
HM updates the randomly selected $w_i$ and $\theta$ for a paratope $P=(w_1,...,w_k,\theta)$ as follows.
\[ w_i=w_i+\Delta w, \theta=\theta+\Delta \theta, \]
where $\Delta w$, $\Delta \theta$ are $-\gamma_w<\Delta w<\gamma_w$, $-1<\Delta \theta <\gamma_\theta$, respectively.

RE makes a crossover of 2 set of $w_i$ for a paratope as shown in Fig.\ref{fig:network_gene}.

\subsubsection{Affinity}
The system calculates the degree of affinities between antibody and antigen by using Eq.(\ref{eq:affinity}) and Eq.(\ref{eq:affinity2}).
\begin{equation}
f(x^p)=\left \{
\begin{array}{l l}
1 & if\ | \sum_{i=1}^{k} w_ix_{i}^{P} -\theta | \geq E_{sim}\\
0 &  otherwise
\end{array}
\right.
\label{eq:affinity}
\end{equation}

\begin{equation}
g(x^p)=\left \{
\begin{array}{l l}
1 & if \ \ f(x^p)=x_{CHD}^p\\
0 &  otherwise
\end{array}
\right.
\label{eq:affinity2}
\end{equation}

Eq.(\ref{eq:affinity3}) calculates the degree of affinity.

\begin{equation}
\label{eq:affinity3}
Affinity=\sum_{p=1}^{tr\_num} g(x^p), 
\end{equation}
where $x^p$ means the $p$th sample in $tr\_num$ training cases and $x_{CHD}^p$ is 1 if an example is a CHD, otherwise 0.

\section{Immunological Memory Cell}
\label{sec:IMC}
\subsection{Clustering Memory Cell}
\begin{figure}[htbp]
\begin{center}
\includegraphics[scale=0.7]{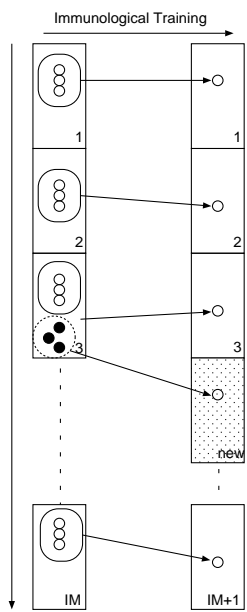}
\caption{A clustering method of memory cells}
\label{fig:CSAIMmodel_clustering}
\end{center}
\end{figure}

Clustering Memory Cells is required to classify the antibodies responding the specified samples. This paper realizes the clustering by allocating the generated antibodies RECSA model into some categories. The initial number of categories is predefined and a new category is created according to training situation. Fig. \ref{fig:CSAIMmodel_clustering} shows the clustering method of memory cells. Similar antibodies crowd around an appropriate point in each category, and then only central antibody of the crowd can become a memory cell. However, we may meet that memory cells can not recognize some of samples in the dataset. In such a case, some new generated antibodies by RECSA model tries to respond to the misclassification of the samples, if the similar antibodies make a crowd.

To find the crowd of similar cases, the system checks whether the Euclidean distance between normalized training sample and its corresponding antibody is smaller than the predetermined parameter $\mu_{\theta}$.

The similarity is measured by the following. Let $\vec{d}=(d_1,\cdots,d_i,\cdots,d_k)$ be the elements of input signal and $\vec{h}=(h_1,\cdots,h_i,\cdots,h_k)$ be the element of antibody.

In order to calculate the distance between the sample and the antibody, the range of sample is changed to that of antibody as follows.
\begin{equation}
\nonumber d_{i}^{'}=d_{i} \times \frac{h_{j}}{d_{j}} (d_{i} \neq  0 \wedge h_{i} \neq 0)
\end{equation}
$d_{j}$ is the min value of element in the input sample.

Then, if the Euclidean distance between $\vec{d^{'}}$ and $\vec{h}$ is smaller than $\mu_{\theta}$, the antibody can respond the sample. In this paper, $\mu_{\theta}$ is the summation of 8 input elements.

\subsection{Training Memory Cell}
Fig.\ref{fig:CSAIMmodel} shows the immune process of CSAIM model. The learning algorithm in CSAIM is as follows.

\begin{center}
%\begin{indentation}{0.5zw}{0.5zw}
\begin{indentation}{0.1cm}{0.1cm}
\begin{breakbox}
\smallskip
\begin{enumerate}
\item Select the elite antibody without improving the degree of affinity during $\tau$ generations.
\item Let $T_{r_{o}}$ be training cases that $g(x^p)$ in Eq.(\ref{eq:affinity2}) equals to $0$. $T_{r_{o}}^{num}$ is the number of training cases in $T_{r_{o}}$.
\item For each training case in $T_{r_{o}}$, calculate the output of the selected individual by using the following equation.
\vspace{-6mm}
\begin{equation}
O=\sum_{i=1}^{k} w_{i} x_{i},
\end{equation}
where $w_{i}$ is a weight for each paratope.
\label{CSAIM_output}
\item Calculate the difference $\delta$ by using output value and $\theta$.
\begin{equation}
  E = \frac{1}{2}\delta^{2} = \frac{1}{2}(\theta_{q} -O)^{2},
\end{equation}
where $q(1\le q \le IM)$ means the cluster of memory cells.
\item Update weights.
\begin{equation}
 w_{i}=w_{i}+\eta \delta x_{i},
\end{equation}
where $\eta$ takes a real value in $[0.1, 1.0]$.
\label{CSAIM_update}
\item The procedure from \ref{CSAIM_output} to \ref{CSAIM_update} is executed for all the cases in $T_{r_{o}}$ till the given number of iterations reaches, $T_{IM}$ or error $E_{min}$ becomes less than the expected value. The good antibodies after training are recorded in the memory.
\end{enumerate}
\end{breakbox}
\end{indentation}
\figcaption{The learning algorithm of CSAIM}
\label{fig:CSAIM}
\end{center}

\section{Experimental Results}
\label{sec:ExperimentalResults}
\subsection{Database Design}
The CHD\_DB is designed to reproduce the original data of the Framingham Heart Study. Requisite information is derived from the reports on six-year follow-up in the Framingham Heart Study \cite{Dawber},\cite{Kannel}.

\begin{figure}
\begin{center}
\includegraphics[scale=0.3]{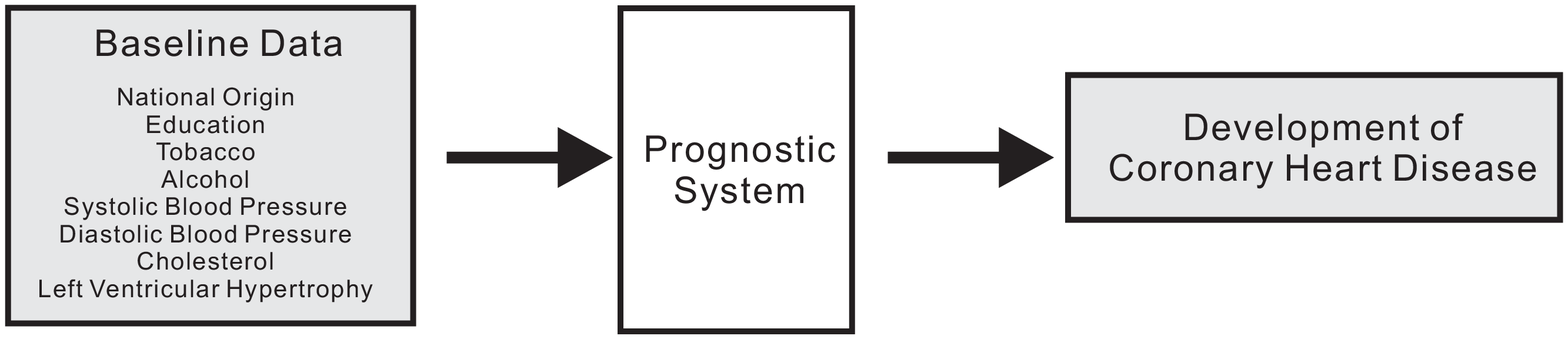}
\caption{Prognostic System for the development of CHD}
\label{fig:CHDflow}
\end{center}
\end{figure}

Table \ref{table:CHDdata} shows the data items of the CHD\_DB. Each of the datasets consists of ten data items: ID, development of CHD, and eight items that were collected from the initial examination (i.e. baseline data). The eight items; ORIGIN, EDUCATE, TOBACCO, ALCHOL, SBP, DBP, TC, and LVH, were examined whether it was associated with the development of CHD over six-year follow-up in the Framingham Heart Study. Using the CHD\_DB, researchers will develop a prognostic system that will discriminate between those who developed CHD (CHD cases) and those who did not (Non-CHD cases) on the basis of the eight data items as shown in Fig.\ref{fig:CHDflow}.

\begin{table*}[htbp]
\begin{center}\caption{Data items of Coronary heart disease database}
\begin{tabular}{ l|l|l} \hline
Data Item & Name & Value \\ \hline
ID & ID & Sequential Value \\
Development of CHD & CHD & 0=No; 1= Yes\\
National Origin & ORIGIN & 0=Native-born; 1=Foreign-born\\
Education & EDUCATE & 0=Grade School; 1=High School, not graduate; 2=High School. graduate; 3=College\\
Tobacco & TOBACCO & 0=Never; 1=Stopped; 2=Cigars or Pipes; 3=Cigarettes($<$20/day); 4=Cigarettes($\leq$20/day)\\
Alcohol & ALCOHOL & Continuous Value(oz/mo)\\
Systolic Blood Pressure & SBP & Continuous Value (mmHg)\\
Diastolic Blood Pressure & DBP & Continuous Value (mmHg)\\
Cholesterol & TC &Continuous Value (mg/dl)\\
Left Ventricular Hypertrophy & LVH & 0=None; 1=Definite or Possible\\ \hline
\end{tabular}
\label{table:CHDdata}
\end{center}
\end{table*}

\begin{figure}
\begin{center}
\includegraphics[scale=0.4]{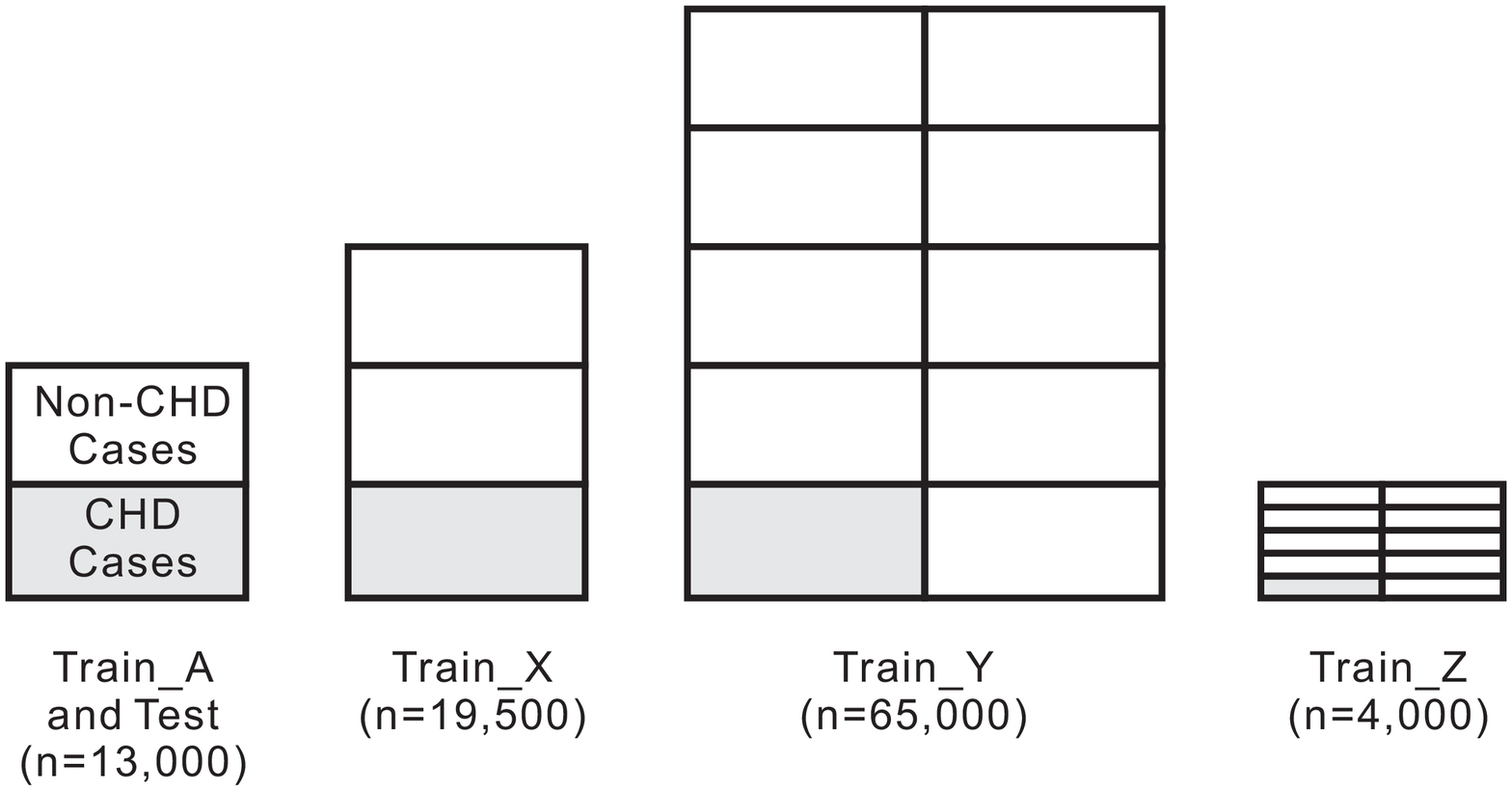}
\caption{Training and test datasets}
\label{fig:CHDdatasets}
\end{center}
\end{figure}

The CHD\_DB consists of four training datasets (Train\_A, X, Y, and Z) and one testing dataset (Test) as shown in Fig.\ref{fig:CHDdatasets}. We previously reported that the developed prognostic system highly depended on the quality of training dataset \cite{Suka-Medinfo}. The small proportion of cases to non-cases in the training dataset might contribute to a poor learning about the cases, and consequently, the developed prognostic system might have lack of ability to identify potential cases from the population at risk. Therefore, the four training datasets are designed to have different proportions of CHD cases to Non-CHD cases. The number of combination of the eight data items is expected to be about six and a half thousand: 2 (ORIGIN) $\times$ 4 (EDUCATE) $\times$ 5 (TOBACCO) $\times$ 3 (ALCOHOL) $\times$ 3 (SBP) $\times$ 3 (DBP) $\times$ 3 (TC) $\times$ 2 (LVH) = 6,480, where the numerical data are counted as three. In the original data of the Framingham Heart Study, the number of complete records was about four thousand and the proportion of CHD cases to Non-CHD cases in men aged 45 or older was about one to nine (i.e. the six-year incidence of CHD was 9.1\%). Therefore, Train\_A, Train\_X, and Train\_Y include 6,500 CHD cases and 6,500 ($\times$ 1), 13,000 ($\times$ 2), and 585,000 ($\times$ 9) Non-CHD cases, respectively, while Train\_Z approximates to the original data both for the number of total records and the proportion of CHD cases to Non-CHD cases. On the other hand, for the testing dataset, we prepared 6,500 CHD cases and 6,500 Non-CHD cases separately from the training datasets.

\subsection{Experimental Results}
In this paper, we examined the classification capability to Train\_A and Test. The parameters in the experiments are defined as follows: $G_{max}=100$, $m=150$, $n=100$, $Q=50$, $HM:RE=1:1$. $-1 \leq r_{w} \leq 1$, $-1 \leq r_{\theta} \leq 1$. $E_{sim}=0.05$, $t=10$, $e=10$ described in Section \ref{sec:IMC}. Furthermore, to improve the calculation speed, we set the following parameters:  the training ratio of memory cells, $\eta=0.1$; the Euclidean distance between a sample and an antibody representing the similarity, $\mu_{\theta}=0.3$; the number of iterations of training memory cells, $t_{IM}=50$; the stop condition of training memory cells, $E_{min}=0.001$; the maximum size of memory cells, $c_{max}^{memory}=\frac{1}{2}n$.

Table \ref{tab:Correctratio} was the comparison result of Train\_A dataset and TEST dataset for the average of 10 runs. The RECSA and CSAIM mean the traditional method and our proposed method, respectively. The classification result shows that our proposed method has better capability power than the traditional method.

\begin{table}[!htbp]
\caption{Correct ratio}
\vspace{-3mm}
\label{tab:Correctratio}
\begin{center}
\begin{tabular}{c|c|c}
\hline
&Train\_A &TEST    \\ \hline
RECSA& 62.3\%& 58.7\%\\
CSAIM& 99.6\%& 99.4\% \\
\hline
\end{tabular}
\end{center}
\end{table}

\begin{figure}
\begin{center}
\includegraphics[scale=0.7]{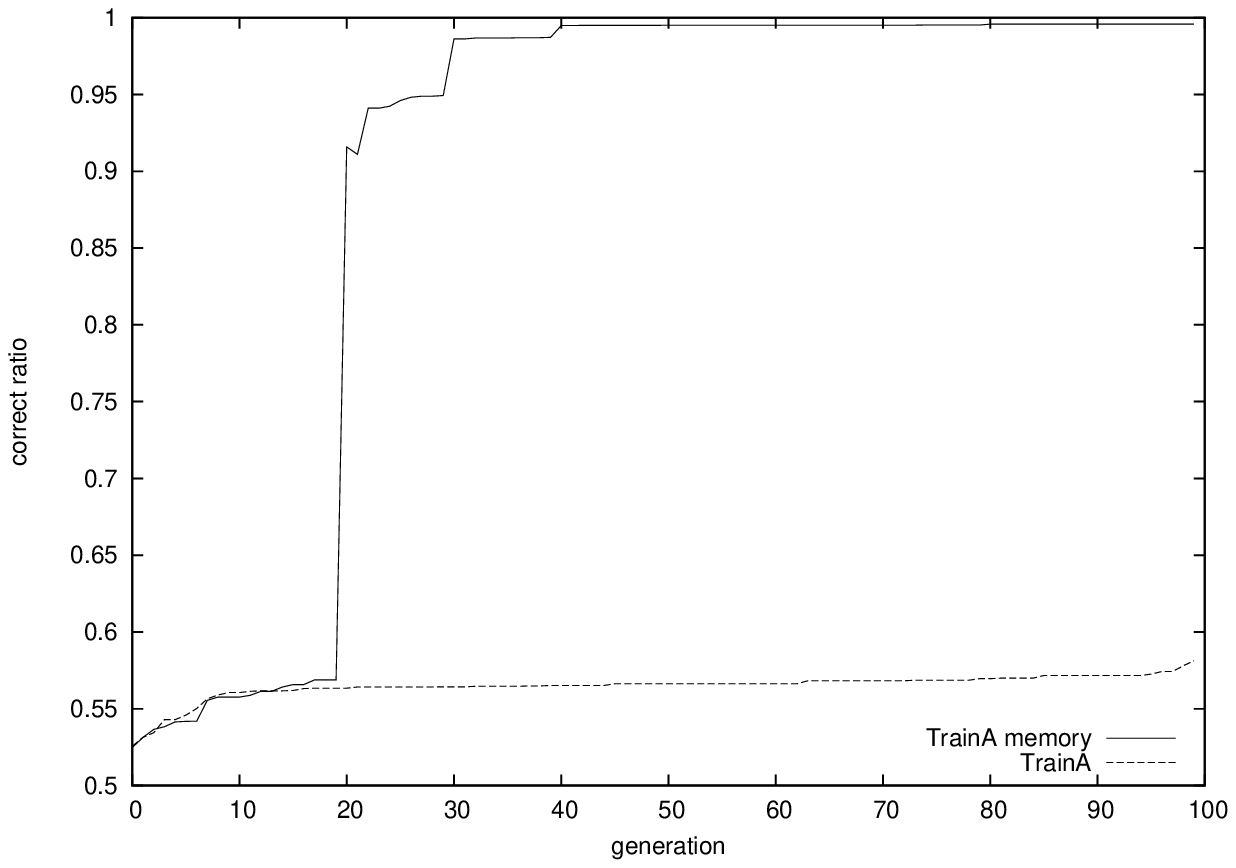}
\caption{Correct ratio of Train\_A}
\label{fig:correct_rate}
\end{center}
\end{figure}

Fig. \ref{fig:correct_rate} shows the correct ratio of training samples at each generation. We can understand that our proposed method can recognize the different patterns even to the patterns which the traditional method cannot classify. Thus, the extraordinary higher capability power can show the good performance of classification of CHD\_DB.

\begin{figure}
\begin{center}
\includegraphics[scale=0.7]{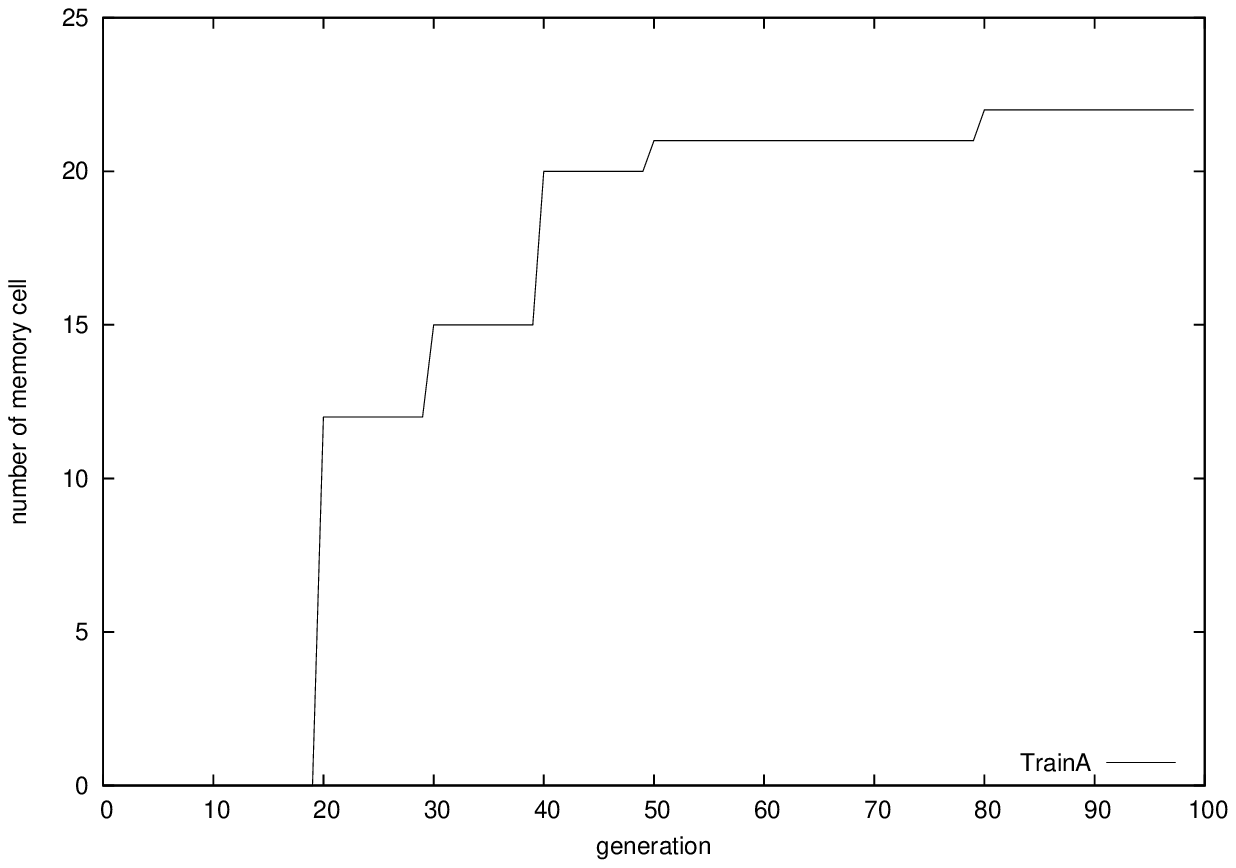}
\caption{The number of Immunological Memory Cells}
\label{fig:NO_IMcell}
\end{center}
\end{figure}

Fig. \ref{fig:NO_IMcell} shows the change of increase of Immunological Memory Cells. From the Fig. \ref{fig:correct_rate} and Fig. \ref{fig:NO_IMcell}, the more memory cells are generated within the maximum size of memory, the classification capability becomes higher.

Furthermore, the calculation speed of our proposed method has been reduced by more than half. Because the system can search an antibody corresponded to the antigen by using memory cells after a period of training, but the system has to generate many antibodies in the initial phase of training.

\section{Conclusive Discussion}
\label{sec:ConclusiveDiscussion}
The clonal selection principle established the idea that only those cells that recognize the antigens are selected to proliferate and differentiate. The method by \cite{Gao}, RECSA, explicitly takes into account the affinity maturation of the immune response by incorporating receptor editing method. In this paper, the immunological memory cells are realized in the RECSA model. A gene structure to classify the Coronary heart disease database (CHD\_DB)\cite{Suka-Data} by using RECSA model was developed. The memory cells in some categories are trained to respond to the specified samples according to the similarities between antigen and antibody. As the computational classification results of Train\_A in CHD\_DB, the classification capability was about 99.6\% correct ratio of test dataset. We will examine the other datasets in CHD\_DB.

% that's all folks
\end{document}